\documentclass[10pt,twocolumn,letterpaper]{article}

\usepackage{iccv}
\usepackage{times}
\usepackage{epsfig}
\usepackage{graphicx}
\usepackage{amsmath}
\usepackage{amssymb}

\usepackage{booktabs}
\usepackage{xcolor}
\usepackage[ruled,vlined]{algorithm2e}
\usepackage{multirow}
\usepackage{wrapfig}
\usepackage{kotex}
\usepackage{caption}
\usepackage{subcaption}
\newcommand{\ve}[1]{\boldsymbol{#1}}
\def \alambicw {\includegraphics[width=0.01\linewidth]{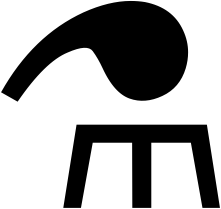}\xspace}
\def \alambic {\includegraphics[width=0.02\linewidth]{figures/alembic-crop.pdf}\xspace}

\definecolor{myblue}{rgb}{0.12, 0.39, 0.60}

\usepackage[pagebackref=true,breaklinks=true,letterpaper=true,colorlinks,bookmarks=false]{hyperref}

\iccvfinalcopy %

\def\iccvPaperID{1311} %

\ificcvfinal\fi

\begin{document}

\title{Rethinking Spatial Dimensions of Vision Transformers}

\author{
Byeongho Heo\textsuperscript{\rm 1} \enspace  
Sangdoo Yun\textsuperscript{\rm 1} \enspace  
Dongyoon Han\textsuperscript{\rm 1} \enspace  
Sanghyuk Chun\textsuperscript{\rm 1} \enspace 
Junsuk Choe\textsuperscript{\rm 2}\thanks{Work done as a research scientist at NAVER AI Lab.} \enspace 
Seong Joon Oh\textsuperscript{\rm 1}\\
\\
\small{\textsuperscript{\rm 1} NAVER AI Lab
\quad
\textsuperscript{\rm 2} Department of Computer Science and Engineering, Sogang University}
}

\maketitle
\ificcvfinal\thispagestyle{empty}\fi

\begin{abstract}
Vision Transformer (ViT) extends the application range of transformers from language processing to computer vision tasks as being an alternative architecture against the existing convolutional neural networks (CNN). Since the transformer-based architecture has been innovative for computer vision modeling, the design convention towards an effective architecture has been less studied yet. From the successful design principles of CNN, we investigate the role of spatial dimension conversion and its effectiveness on transformer-based architecture. We particularly attend to the dimension reduction principle of CNNs; as the depth increases, a conventional CNN increases channel dimension and decreases spatial dimensions. We empirically show that such a spatial dimension reduction is beneficial to a transformer architecture as well, and propose a novel Pooling-based Vision Transformer (PiT) upon the original ViT model. We show that PiT achieves the improved model capability and generalization performance against ViT. Throughout the extensive experiments, we further show PiT outperforms the baseline on several tasks such as image classification, object detection, and robustness evaluation. Source codes and ImageNet models are available at {\url{https://github.com/naver-ai/pit}}.
\end{abstract}

\section{Introduction}

The architectures based on the self-attention mechanism have achieved great success in the field of Natural Language Processing (NLP)~\cite{vaswani2017attention}.
There have been attempts to utilize the self-attention mechanism in computer vision.
Non-local networks~\cite{wang2018nonlocal} and DETR~\cite{carion2020detr} are representative works, showing that the self-attention mechanism is also effective in video classification and object detection tasks, respectively.
Recently, Vision Transformer (ViT)~\cite{vit}, a transformer architecture consisting of self-attention layers, has been proposed to compete with ResNet~\cite{he2016resnet}, and shows that it can achieve the best performance without convolution operation on ImageNet~\cite{deng2009imagenet}.
As a result, a new direction of network architectures based on self-attention mechanism, not convolution operation, has emerged in computer vision.

ViT is quite different from convolutional neural networks (CNN). 
Input images are divided into 16$\times$16 patches and fed to the transformer network; except for the first embedding layer, there is no convolution operation in ViT, and the position interactions occur only through the self-attention layers.
While CNNs have restricted spatial interactions, ViT allows all the positions in an image to interact through transformer layers. 
Although ViT is an innovative architecture and has proven its powerful image recognition ability, it follows the transformer architecture in NLP~\cite{vaswani2017attention} without any changes.
Some essential design principles of CNNs, which have proved to be effective in the computer vision domain over the past decade, are not sufficiently reflected. We thus revisit the design principles of CNN architectures and investigate their efficacy when applied to ViT architectures.

CNNs start with a feature of large spatial sizes and a small channel size and gradually increase the channel size while decreasing the spatial size.
This dimension conversion is indispensable due to the layer called spatial pooling.
Modern CNN architectures, including AlexNet~\cite{krizhevsky2012alexnet}, ResNet~\cite{he2016resnet}, and EfficientNet~\cite{tan2019efficientnet}, follow this design principle.
The pooling layer is deeply related to the receptive field size of each layer.
Some studies~\cite{cohen2016inductive,radenovic2018gempooling,cohen2016expressive} show that the pooling layer contributes to the expressiveness and generalization performance of the network.
However, unlike the CNNs, ViT does not use a pooling layer and uses the same spatial dimension for all layers.

\begin{figure*}[t]
	\centering
    \includegraphics[width=1.0\linewidth]{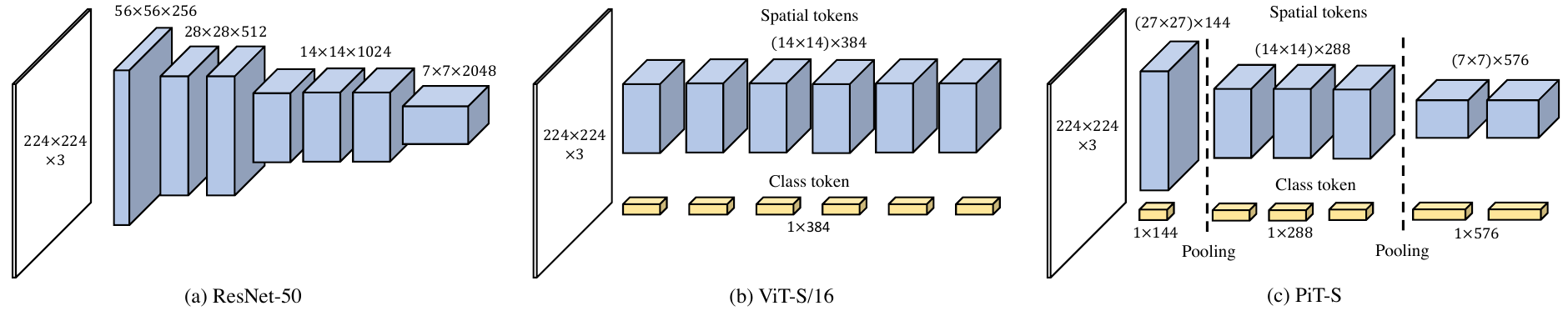}
  \caption{\textbf{Schematic illustration of dimension configurations of networks.} We visualize ResNet50~\cite{he2016resnet}, Vision Transformer (ViT)~\cite{vit}, and our Pooling-based Vision Transformer (PiT); (a) ResNet50 gradually downsamples the features from the input to the output; (b) ViT does not change the spatial dimensions; (c) PiT involves ResNet style spatial dimension into ViT.}
\label{fig:dimension_difference}
\vspace{-0.2cm}
\end{figure*}

First, we verify the advantages of dimensions configurations on CNNs.
Our experiments show that ResNet-style dimensions improve the model capability and generalization performance of ResNet.
To extend the advantages to ViT, we propose a Pooling-based Vision Transformer (PiT).
PiT is a transformer architecture combined with a newly designed pooling layer. 
It enables the spatial size reduction in the ViT structure as in ResNet.
We also investigate the benefits of PiT compared to ViT and confirm that ResNet-style dimension setting also improves the performance of ViT.
Finally, to analyze the effect of PiT compared to ViT, we analyze the attention matrix of transformer block with entropy and average distance measure.
The analysis shows the attention patterns inside layers of ViT and PiT, and helps to understand the inner mechanism of ViT and PiT.

We verify that PiT improves performances over ViT on various tasks. 
On ImageNet classification, PiT and outperforms ViT at various scales and training environments.
Additionally, we have compared the performance of PiT with various convolutional architectures and have specified the scale at which the transformer architecture outperforms the CNN.
We further measure the performance of PiT as a backbone for object detection. ViT- and PiT-based deformable DETR~\cite{zhu2020deformable} are trained on the COCO 2017 dataset~\cite{lin2014coco} and the result shows that PiT is even better than ViT as a backbone architecture for a task other than image classification.
Finally, we verify the performance of PiT in various environments through the robustness benchmark.
\section{Related works}

\subsection{Dimension configuration of CNN}

Dimension conversion can be found in AlexNet~\cite{krizhevsky2012alexnet}, which is one of the earliest convolutional networks in computer vision.
AlexNet uses three max-pooling layers. In the max-pooling layer, the spatial size of the feature is reduced by half, and the channel size is increased by the convolution after the max-pooling.
VGGnet~\cite{simonyan2014vgg} uses 5 spatial resolutions using 5 max-pooling. In the pooling layer, the spatial size is reduced by half and the channel size is doubled.
GoogLeNet~\cite{szegedy2015googlenet} also used the pooling layer. 
ResNet~\cite{he2016resnet} performed spatial size reduction using the convolution layer of stride 2 instead of max pooling.
It is an improvement in the spatial reduction method.
The convolution layer of stride 2 is also used as a pooling method in recent architectures (EfficietNet~\cite{tan2019efficientnet}, MobileNet~\cite{sandler2018mobilenetv2,howard2019mobilev3}).
PyramidNet~\cite{han2017pyramid} pointed out that the channel increase occurs only in the pooling layer and proposed a method to gradually increase the channel size in layers other than the pooling layer.
ReXNet~\cite{han2020rexnet} reported that the channel configuration of the network has a significant influence on the network performance.
In summary, most convolution networks use a dimension configuration with spatial reduction.

\begin{figure*}
     \centering
     \begin{subfigure}[b]{0.33\textwidth}
         \centering
         \includegraphics[width=\textwidth]{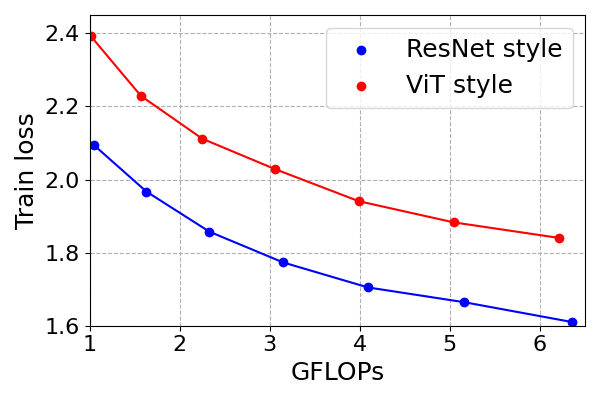}
         \caption{Model capability}
         \label{fig:rn_flops_trainloss}
     \end{subfigure}
     \hfill
     \begin{subfigure}[b]{0.33\textwidth}
         \centering
         \includegraphics[width=\textwidth]{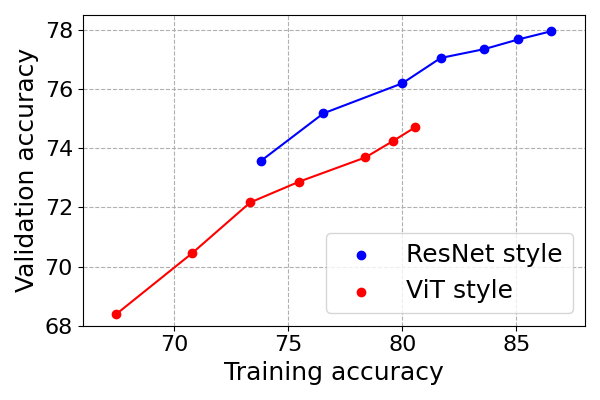}
         \caption{Generalization performance}
         \label{fig:rn_trainacc_valacc}
     \end{subfigure}
     \hfill
     \begin{subfigure}[b]{0.33\textwidth}
         \centering
         \includegraphics[width=\textwidth]{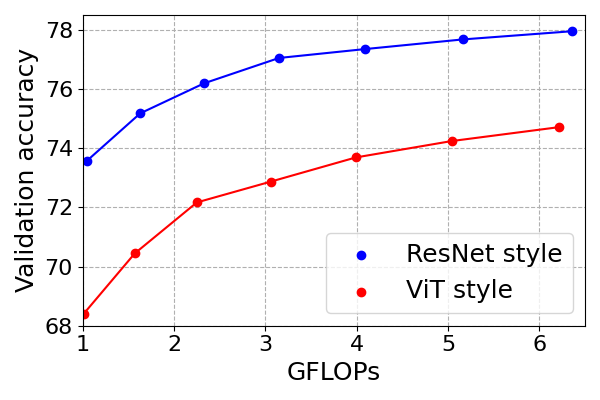}
         \caption{Model performance}
         \label{fig:rn_gflops_valacc}
     \end{subfigure}
        \caption{\textbf{Effects of the spatial dimensions in ResNet50~\cite{he2016resnet}.} We verify the effect of the spatial dimension with ResNet50. As shown in the figures, ResNet-style is better than ViT-style in the model capability, generalization performance, and model performance.}
        \label{fig:resnet_pooling_experiment}
\end{figure*}

\subsection{Self-attention mechanism}

Transformer architecture~\cite{vaswani2017attention} significantly increased the performance of the NLP task with the self-attention mechanism. 
Funnel Transformer~\cite{dai2020funnel} improves the transformer architecture by reducing tokens by a pooling layer and skip-connection.
However, because of the basic difference between the architecture of NLP and computer vision, the method of applying to pool is different from our method.
Some studies are conducted to utilize the transformer architecture to the backbone network for computer vision tasks.
Non-local network~\cite{wang2018nonlocal} adds a few self-attention layers to CNN backbone, and it shows that the self-attention mechanism can be used in CNN.
\cite{ramachandran2019stand} replaced $3\times3$ convolution of ResNet to local self-attention layer.
\cite{wang2020axial} used an attention layer for each spatial axis.
\cite{bello2021lambdanetworks} enables self-attention of the entire spatial map by reducing the computation of the attention mechanism.
Most of these methods replace 3x3 convolution with self-attention or adds a few self-attention layers.
Therefore, the basic structure of ResNet is inherited, that is, it has the convolution of stride 2 as ResNet, resulting in a network having a dimension configuration of ResNet.

Only the vision transformer uses a structure that uses the same spatial size in all layers.
Although ViT did not follow the conventions of ResNet, it contains many valuable new components in the network architecture.
In ViT, layer normalization is applied for each spatial token. 
Therefore, layer normalization of ViT is closer to positional normalization~\cite{pono} than a layer norm of convolutional neural network~\cite{ba2016layernorm,wu2018groupnorm}.
Although it overlaps with the lambda network~\cite{bello2021lambdanetworks}, it is not common to use global attention through all blocks of the network.
The use of class tokens instead of global average pooling is also new, and it has been reported that separating tokens increases the efficiency of distillation~\cite{deit}.
In addition, the layer configuration, the skip-connection position, and the normalization position of the Transformer are also different from ResNet.
Therefore, our study gives a direction to the new architecture.

\section{Revisiting spatial dimensions}

In order to introduce dimension conversion to ViT, we investigate spatial dimensions in network architectures.
First, we verify the benefits of dimension configuration in ResNet architecture.
Although dimension conversion has been widely used for most convolutional architectures, its effectiveness is rarely verified.
Based on the findings, we propose a Pooling-based Vision Transformer (PiT) that applies the ResNet-style dimension to ViT.
We propose a new pooling layer for transformer architecture and design ViT with the new pooling layer (PiT).
With PiT models, we verify whether the ResNet-style dimension brings advantages to ViT.
In addition, we analyze the attention matrix of the self-attention block of ViT to investigate the effect of PiT in the transformer mechanism.
Finally, we introduce PiT architectures corresponding to various scales of ViT.

\begin{figure*}
     \centering
     \begin{subfigure}[b]{0.33\textwidth}
         \centering
         \includegraphics[width=\textwidth]{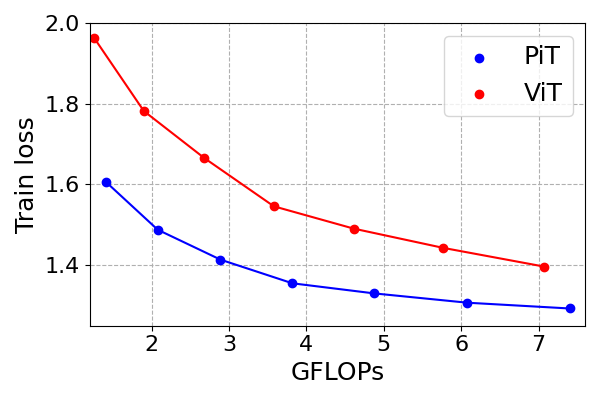}
         \caption{Model capability}
         \label{fig:vit_flops_trainloss}
     \end{subfigure}
     \hfill
     \begin{subfigure}[b]{0.33\textwidth}
         \centering
         \includegraphics[width=\textwidth]{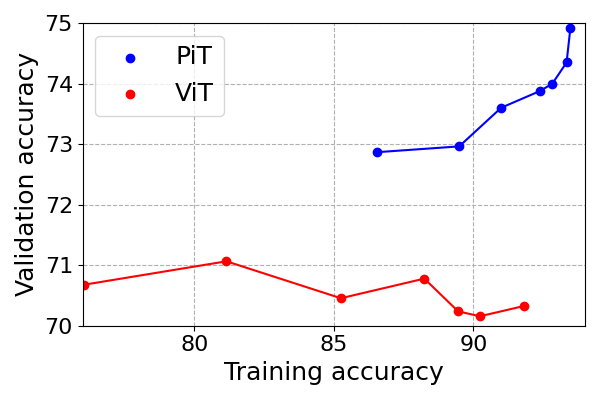}
         \caption{Generalization performance}
         \label{fig:vit_trainacc_valacc}
     \end{subfigure}
     \hfill
     \begin{subfigure}[b]{0.33\textwidth}
         \centering
         \includegraphics[width=\textwidth]{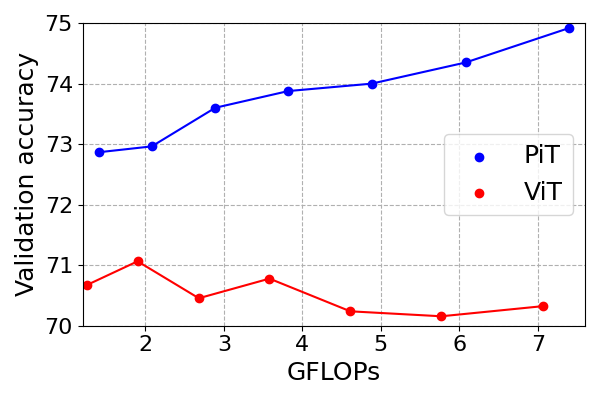}
         \caption{Model performance}
         \label{fig:vit_gflops_valacc}
     \end{subfigure}
        \caption{\small \textbf{Effects of the spatial dimensions in vision transformer (ViT)~\cite{vit}.} We compare our Pooling-based Vision Transformer (PiT) with original ViT at various aspects. PiT outperforms ViT in capability, generalization performance, and model performance.}
        \label{fig:vit_pooling_experiment}
\end{figure*}

\subsection{Dimension setting of CNN}

As shown in Figure~\ref{fig:dimension_difference} (a), most convolutional architectures reduce the spatial dimension while increases the channel dimension.
In ResNet50, a stem layer reduces the spatial size of an image to $56\times56$.
After several layer blocks, Convolution layers with stride 2 reduce the spatial dimension by half and double the channel dimension.
The spatial reduction using a convolution layer with stride 2 is a frequently used method in recent architectures~\cite{tan2019efficientnet,sandler2018mobilenetv2,howard2019mobilev3,han2020rexnet}.
We conduct an experiment to analyze the performance difference according to the presence or absence of the spatial reduction layer in a convolutional architecture.
ResNet50, one of the most widely used networks in ImageNet, is used for architecture and is trained over 100 epochs without complex training techniques.
For ResNet with ViT style dimension, we use the stem layer of ViT to reduce the feature to $14\times14$ spatial dimensions while reducing the spatial information loss in the stem layer.
We also remove the spatial reduction layers of ResNet to maintain the initial feature dimensions for all layers like ViT.
We measured the performance for several sizes by changing the channel size of ResNet.

First, we measured the relation between FLOPs and training loss of ResNet with ResNet-style or ViT-style dimension configuration.
As shown in Figure~\ref{fig:resnet_pooling_experiment} (a), ResNet (ResNet-style) shows lower training loss over the same computation costs (FLOPs).
It implies that ResNet-style dimensions increase the capability of architecture.
Next, we analyzed the relation between training and validation accuracy, which represents the generalization performance of architecture.
As shown in Figure~\ref{fig:resnet_pooling_experiment} (b), ResNet (ResNet-style) achieves higher validation accuracy than ResNet  (ViT-style).
Therefore, ResNet-style dimension configuration is also helpful for generalization performance.
In summary, ResNet-style dimension improves the model capability and generalization performance of the architecture and consequently brings a significant improvement in validation accuracy as shown in Figure~\ref{fig:resnet_pooling_experiment} (c).

\begin{figure}[t]
	\centering
    \includegraphics[width=1.0\columnwidth]{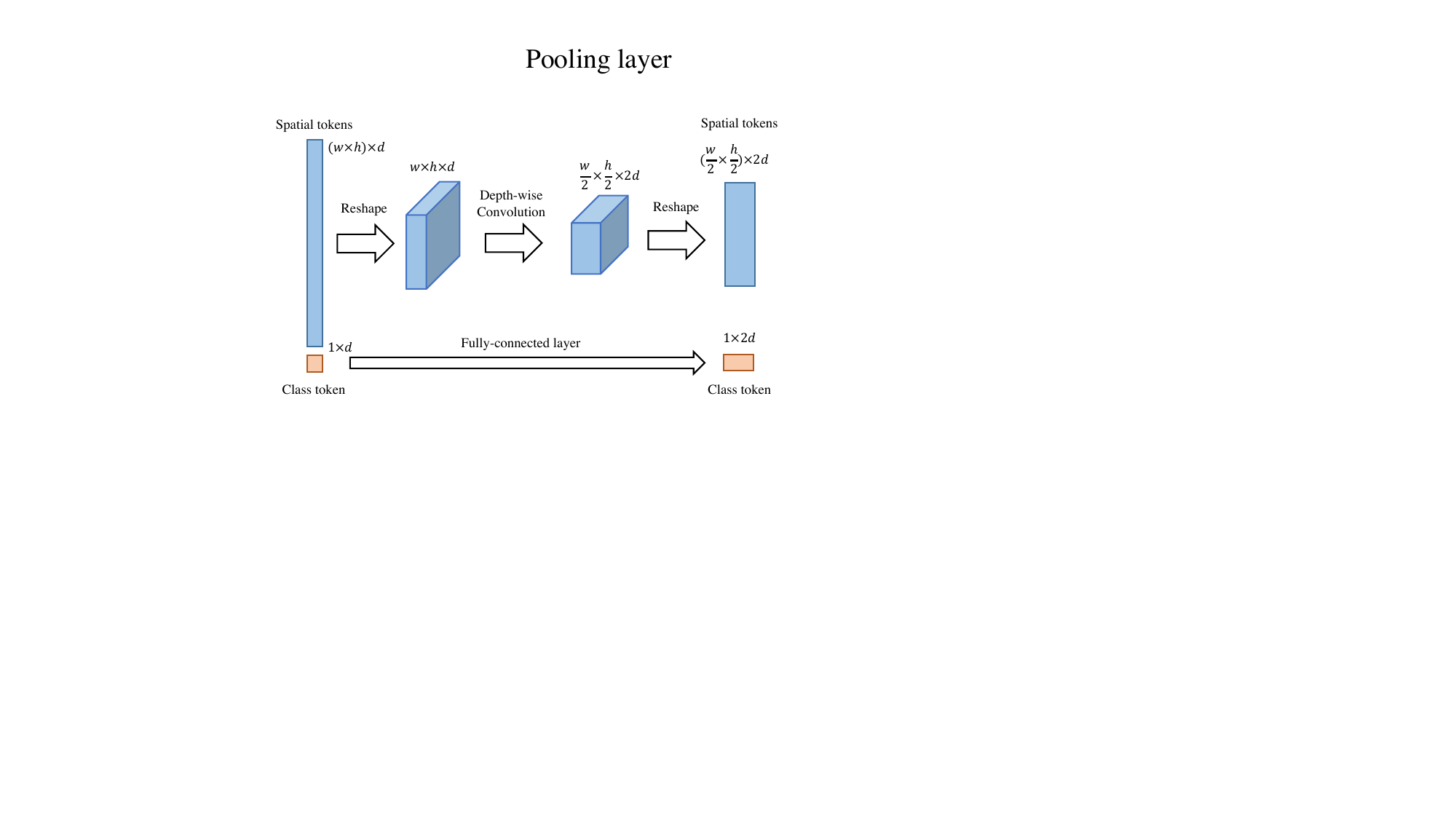}
  \caption{\textbf{Pooling layer of PiT architecture.} PiT uses the pooling layer based on depth-wise convolution to achieve channel multiplication and spatial reduction with small parameters.}
\label{fig:pooling_layer}
\end{figure}

\subsection{Pooling-based Vision Transformer (PiT)}

Vision Transformer (ViT) performs network operations based on self-attention, not convolution operations.
In the self-attention mechanism, the similarity between all locations is used for spatial interaction. 
Figure~\ref{fig:dimension_difference} (b) shows the dimension structure of this ViT.
Similar to the stem layer of CNN, ViT divides the image by patch at the first embedding layer and embedding it to tokens.
Basically, the structure does not include a spatial reduction layer and keeps the same number of spatial tokens overall layer of the network.
Although the self-attention operation is not limited by spatial distance, the size of the spatial area participating in attention is affected by the spatial size of the feature.
Therefore, in order to adjust the dimension configuration like ResNet, a spatial reduction layer is also required in ViT.

To utilize the advantages of the dimension configuration to ViT, we propose a new architecture called Pooling-based Vision Transformer (PiT).
First, we designed a pooling layer for ViT.
Our pooling layer is shown in Figure~\ref{fig:pooling_layer}.
Since ViT handles neuron responses in the form of 2D-matrix rather than 3D-tensor, the pooling layer should separate spatial tokens and reshape them into 3D-tensor with spatial structure.
After reshaping, spatial size reduction and channel increase are performed by depth-wise convolution.
And, the responses are reshaped into a 2D matrix for the computation of transformer blocks.
In ViT, there are parts that do not correspond to the spatial structure, such as a class token or distillation token~\cite{deit}.
For these parts, the pooling layer uses an additional fully-connected layer to adjust the channel size to match the spatial tokens.
Our pooling layer enables spatial reduction on ViT and is used for our PiT architecture as shown in Figure~\ref{fig:dimension_difference} (c).
PiT includes two pooling layers which make three spatial scales.

Using PiT architecture, we performed an experiment to verify the effect of PiT compared to ViT.
The experiment setting is the same as the ResNet experiment.
Figure~\ref{fig:vit_pooling_experiment} (a) represents the model capability of ViT and PiT.
At the same computation cost, PiT has a lower train loss than ViT.
Using the spatial reduction layers in ViT also improves the capability of architecture.
The comparison between training accuracy and validation accuracy shows a significant difference.
As shown in Figure~\ref{fig:vit_pooling_experiment} (b), ViT does not improve validation accuracy even if training accuracy increases. 
On the other hand, in the case of PiT, validation accuracy increases as training accuracy increases.
The big difference in generalization performance causes the performance difference between PiT and ViT as shown in Figure~\ref{fig:vit_pooling_experiment} (c).
The phenomenon that ViT does not increase performance even when FLOPs increase in ImageNet is reported in ViT paper~\cite{vit}.
In the training data of ImageNet scale, ViT shows poor generalization performance, and PiT alleviates this.
So, we believe that the spatial reduction layer is also necessary for the generalization of ViT.
Using the training trick is a way to improve the generalization performance of ViT in ImageNet.
The combination of training tricks and PiT is covered in the experiment section.

\subsection{Attention analysis}
We analyze the transformer networks with measures on attention matrix~\cite{vig2019analyzing}.
We denotes $\alpha_{i,j}$ as $(i,j)$ component of attention matrix $A\in\mathbb{R}^{M \times N}$. Note that attention values after soft-max layer is used, i.e. $\sum_i \alpha_{i,j} = 1$. 
The attention entropy is defined as 
\begin{equation}
    \text{Entropy} = - \frac{1}{N} \sum_j^N \sum_i \alpha_{i,j} \log \alpha_{i,j}.
\end{equation}
The entropy shows the spread and concentration degree of an attention interaction.
A small entropy indicates a concentrated interaction, and a large entropy indicates a spread interaction.
We also measure an attention distance,
\begin{equation}
    \text{Distance} = \frac{1}{N} \sum_j^N \sum_i \alpha_{i,j} \|\ve{p}_i - \ve{p}_j\|_1.
\end{equation}
$\ve{p}_i$ represents relative spatial location of $i$-th token $(x_i/W, y_i/H)$ for feature map $F\in\mathbb{R}^{H\times W\times C}$. 
So, the attention distance shows a relative ratio compared to the overall feature size, which enables comparison between the different sizes of features.
We analyze transformer-based models (ViT-S~\cite{deit} and PiT-S) and values are measured overall validation images and are averaged over all heads of each layer.
Our analysis is only conducted for the spatial tokens rather than the class token following the previous study~\cite{vig2019analyzing}.
We also skip the attention of the last transformer block since the spatial tokens of the last attention are independent of the network outputs.

\begin{figure}
     \centering
     \begin{subfigure}[b]{0.45\textwidth}
         \centering
         \includegraphics[width=1.0\textwidth]{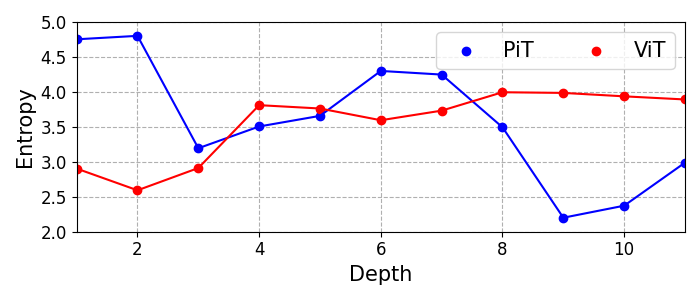}
         \caption{Attention entropy}
         \label{fig:vit_flops_trainloss}
     \end{subfigure}
     
     \begin{subfigure}[b]{0.45\textwidth}
         \centering
         \includegraphics[width=1.0\textwidth]{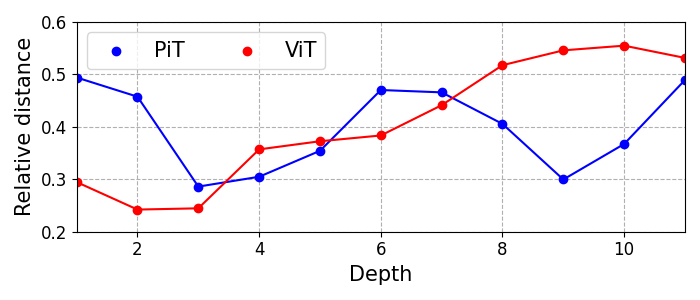}
         \caption{Spatial distance of interaction}
         \label{fig:vit_trainacc_valacc}
     \end{subfigure}
    \caption{\small \textbf{Attention analysis.} We investigate the attention matrix of the self-attention layer. Figure (a) shows the entropy and figure (b) shows the interaction distance. PiT increases the entropy and the distance in shallow layers and decreases in deep layers.}
    \label{fig:attention_analysis}
\end{figure}

The results are shown in Figure~\ref{fig:attention_analysis}.
In ViT, the entropy and the distance increase as the layer become deeper.
It implies that the interaction of ViT is concentrated to close tokens at the shallow layers and the interaction is spread  in a wide range of tokens at the deep layers.
The entropy and distance pattern of ViT is similar to the pattern of transformer in the language domain~\cite{vig2019analyzing}.
PiT changes the patterns with the spatial dimension setting.
At shallow layers (1-2 layers), large spatial size increases the entropy and distance.
On the other hand,  the entropy and distance are decreased at deep layers (9-11 layers) due to the small spatial size.
In short, the pooling layer of PiT spreads the interaction in the shallow layers and concentrates the interaction in the deep layers.
In contrast to discrete word inputs of the language domain, the vision domain uses image-patch inputs which require pre-processing operations such as filtering, contrast, and brightness calibration.
In shallow layers, the spread interaction of PiT is close to the pre-processing than the concentrated interaction of ViT.
Also, compared to language models, image recognition has relatively low output complexity.
So, in deep layers, concentrated interaction might be enough.
There are significant differences between the vision and the language domain, and we believe that the attention of PiT is suitable for image recognition backbone.

\begin{table}[t]
\centering
\small
\begin{tabular}{@{}cccccc@{}}
\toprule
Network                 & \begin{tabular}[c]{@{}c@{}}Spatial\\ size\end{tabular} & \begin{tabular}[c]{@{}c@{}}\# of\\ blocks\end{tabular} & \begin{tabular}[c]{@{}c@{}}\# of\\ heads\end{tabular} & \begin{tabular}[c]{@{}c@{}}Channel\\ size\end{tabular} & FLOPs    \\ \midrule
ViT-Ti~\cite{deit}      & 14 x 14      & 12           & 3           & 192      & 1.3B                  \\ \midrule
\multirow{3}{*}{PiT-Ti} & 27 x 27      & 2            & 2           & 64       & \multirow{3}{*}{0.7B} \\
                        & 14 x 14      & 6            & 4           & 128      &                       \\
                        & 7 x 7        & 4            & 8           & 256      &                       \\ \midrule
\multirow{3}{*}{PiT-XS} & 27 x 27      & 2            & 2           & 96       & \multirow{3}{*}{1.4B} \\
                        & 14 x 14      & 6            & 4           & 192      &                       \\
                        & 7 x 7        & 4            & 8           & 384      &                       \\ \midrule \midrule
ViT-S~\cite{deit}       & 14 x 14      & 12           & 6           & 384      & 4.6B                  \\ \midrule
\multirow{3}{*}{PiT-S}  & 27 x 27      & 2            & 3           & 144      & \multirow{3}{*}{2.9B} \\
                        & 14 x 14      & 6            & 6           & 288      &                       \\
                        & 7 x 7        & 4            & 12          & 576      &                       \\ \midrule \midrule
ViT-B~\cite{vit}        & 14 x 14      & 12           & 12          & 768      & 17.6B                 \\ \midrule
\multirow{3}{*}{PiT-B}  & 31 x 31      & 3            & 4           & 256      & \multirow{3}{*}{12.5B}\\
                        & 16 x 16      & 6            & 8           & 512      &                       \\
                        & 8 x 8        & 4            & 16          & 1024     &                       \\ \bottomrule
\end{tabular}
\caption{\small \textbf{Architecture configuration.} The table shows spatial sizes, number of blocks, number of heads, channel size, and FLOPs of ViT and PiT. The structure of PiT is designed to be as similar as possible to ViT and to have less GPU latency.}
\label{table:pit_architecture}
\vspace{-1em}
\end{table}

\begin{table*}[t]
\small
\centering
\begin{tabular}{@{}rccccccc@{}}
\toprule
Architecture & FLOPs  & \begin{tabular}[c]{@{}c@{}}\# of \\ params\end{tabular} & \begin{tabular}[c]{@{}c@{}}Throughput \\ (imgs/sec)\end{tabular} & Vanilla & +CutMix~\cite{yun2019cutmix} & +DeiT~\cite{deit} & +Distill\alambicw~\cite{deit} \\ \midrule
ViT-Ti~\cite{deit} & 1.3 B  & 5.7 M     & 2564   & 68.7\% & 68.5\%    & 72.2\%  & 74.5\%    \\
PiT-Ti             & 0.7 B & 4.9 M      & 3030   & 71.3\% & 72.6\%    & 73.0\%  & 74.6\%        \\
PiT-XS             & 1.4 B  & 10.6 M    & 2128   & 72.4\% & 76.8\%    & 78.1\%  & 79.1\%        \\ \midrule
ViT-S~\cite{deit}  & 4.6 B  & 22.1 M    & 980    & 68.7\% & 76.5\%    & 79.8\%  & 81.2\%        \\
PiT-S              & 2.9 B  & 23.5 M    & 1266   & 73.3\% & 79.0\%    & 80.9\%  & 81.9\%        \\ \midrule
ViT-B~\cite{vit}   & 17.6 B & 86.6 M    & 303    & 69.3\% & 75.3\%    & 81.8\%  & 83.4\%        \\
PiT-B              & 12.5 B & 73.8 M    & 348    & 76.1\% & 79.9\%    & 82.0\%  & 84.0\%        \\ \bottomrule
\end{tabular}
\caption{\small \textbf{ImageNet performance comparison with ViT.} We compare the performances of ViT and PiT with some training techniques on ImageNet dataset. PiT shows better performance with low computation compared to ViT.}
\label{table:vit_pit_imagenet}
\vspace{-1em}
\end{table*}

\subsection{Architecture design}

The architectures proposed in ViT paper~\cite{vit} aimed at datasets larger than ImageNet.
These architectures (ViT-Large, ViT-Huge) have an extremely large scale than general ImageNet networks, so it is not easy to compare them with other networks.
So, following the previous study~\cite{deit} of Vision Transformer on ImageNet, we design the PiT at a scale similar to the small-scale ViT architectures (ViT-Base, ViT-Small, ViT-Tiny).
In the DeiT paper~\cite{deit}, ViT-Small and ViT-Tiny are named DeiT-S and DeiT-Ti, but to avoid confusion due to the model name change, we use ViT for all models.
Corresponding to the three scales of ViT (tiny, small, and base), we design four scales of PiT (tiny, extra small, small, and base). Detail architectures are described in Table~\ref{table:pit_architecture}. 
For convenience, we abbreviate the model names: Tiny - Ti, eXtra Small - XS, Small - S, Base - B
FLOPs and spatial size were measured based on $224\times224$ image.
Since PiT uses a larger spatial size than ViT, we reduce the stride size of the embedding layer to 8, while patch-size is 16 as ViT.
Two pooling layers are used for PiT, and the channel increase is implemented as increasing the number of heads of multi-head attention.
We design PiT to have a similar depth to ViT, and adjust the channels and the heads to have smaller FLOPs, parameter size, and GPU latency than those of ViT.
We clarify that PiT is not designed with large-scale parameter search such as NAS~\cite{liu2018darts,cai2018proxylessnas}, so PiT can be further improved through a network architecture search.

\section{Experiments}

We verified the performance of PiT through various experiments.
First, we compared PiT at various scales with ViT in various training environments of ImageNet training.
And, we extended the ImageNet comparison to architectures other than Transformer.
In particular, we focus on the comparison of the performance of ResNet and PiT, and investigate whether PiT can beat ResNet.
We also applied PiT to an object detector based on deformable DETR~\cite{zhu2020deformable}, and compared the performance as a backbone architecture for object detection.
To analyze PiT in various views, we evaluated the performance of PiT on robustness benchmarks.

\begin{table}[t]
\centering
\small
\begin{tabular}{@{}cccc@{}}
\toprule
Network & \begin{tabular}[c]{@{}c@{}}\# of \\ params\end{tabular}  & \begin{tabular}[c]{@{}c@{}}Throughput \\ (imgs/sec)\end{tabular} & Accuracy \\ \midrule
ResNet18~\cite{he2016resnet,yun2021relabel}  & 11.7M    & 4545      & 72.5\%     \\
MobileNetV2~\cite{sandler2018mobilenetv2}  & 3.5M    & 3846      & 72.0\%     \\
MobileNetV3~\cite{howard2019mobilev3}  & 5.5M    & 3846      & 75.2\%     \\
EfficientNet-B0~\cite{tan2019efficientnet}  & 5.3M    & 2857      & 77.1\%     \\
ViT-Ti~\cite{deit}            & 5.7M     & 2564      & 72.2\%     \\
\textbf{PiT-Ti}    & 4.9M     & 3030      & 73.0\%     \\
ViT-Ti\alambic~\cite{deit}            & 5.7M     & 2564      & 74.5\%     \\
\textbf{PiT-Ti\alambic}    & 4.9M     & 3030      & 74.6\%     \\ \midrule
ResNet34~\cite{he2016resnet,rw2019timm}  & 21.8M    & 2631      & 75.1\%     \\
ResNet34D~\cite{He_2019_bag,rw2019timm}  & 21.8M    & 2325      & 77.1\%     \\
EfficientNet-B1~\cite{tan2019efficientnet}  & 7.8M    & 1754      & 79.1\%     \\
\textbf{PiT-XS}    & 10.6M     & 2128     & 78.1\%     \\
\textbf{PiT-XS\alambic}    & 10.6M     & 2128     & 79.1\%     \\ \midrule
ResNet50~\cite{he2016resnet,yun2021relabel}  & 25.6M    & 1266      & 80.2\%     \\
ResNet101~\cite{he2016resnet,yun2021relabel}  & 44.6M    & 757      & 81.6\%     \\
ResNet50D~\cite{He_2019_bag,rw2019timm}  & 25.6M    & 1176      & 80.5\%     \\
EfficientNet-B2~\cite{tan2019efficientnet}  & 9.2M    & 1333      & 80.1\%     \\
EfficientNet-B3~\cite{tan2019efficientnet}  & 12.2M    & 806      & 81.6\%     \\
RegNetY-4GF~\cite{radosavovic2020regnet}  & 20.6M    & 1136      & 79.4\%     \\
ResNeSt50~\cite{zhang2020resnest}  & 27.5M    & 877      & 81.1\%     \\
ViT-S~\cite{deit}     & 22.1M     & 980      & 79.8\%     \\
\textbf{PiT-S}     & 23.5M     & 1266     & 80.9\%     \\
ViT-S\alambic~\cite{deit}     & 22.1M     & 980      & 81.2\%     \\
\textbf{PiT-S\alambic}     & 23.5M     & 1266     & 81.9\%     \\ \midrule
ResNet152~\cite{he2016resnet,yun2021relabel}  & 60.2M    & 420      & 81.9\%     \\
ResNet101D~\cite{He_2019_bag,rw2019timm}  & 44.6M    & 354      & 83.0\%     \\
ResNet152D~\cite{He_2019_bag,rw2019timm}  & 60.2M    & 251      & 83.7\%     \\
EfficientNet-B4~\cite{tan2019efficientnet}  & 19.3M    & 368      & 82.9\%     \\
RegNetY-16GF~\cite{radosavovic2020regnet}  & 83.6M    & 352      & 80.4\%     \\
ResNeSt101~\cite{zhang2020resnest}  & 48.3M    & 398      & 83.0\%     \\
ViT-B~\cite{vit,deit}     & 86.6M     & 303      & 81.8\%     \\
\textbf{PiT-B}     & 73.8M     & 348      & 82.0\%     \\
ViT-B\alambic~\cite{vit,deit}     & 86.6M     & 303      & 83.4\%     \\
\textbf{PiT-B\alambic}     & 73.8M     & 348      & 84.0\%     \\ \bottomrule
\end{tabular}
\caption{\textbf{ImageNet performance.} We compare our PiT-(Ti, XS, S, and B) models with the counterparts which have a similar number of parameters. \alambic means a model trained with distillation~\cite{deit}.}
\label{table:imagenet_models}
\vspace{-1em}
\end{table}

\subsection{ImageNet classification}

We compared the performance of PiT models of Table~\ref{table:pit_architecture} with corresponding ViT models.
To clarify the computation time and size of the network, we measured FLOPs, the number of parameters, and GPU throughput (images/sec) of each network.
The GPU throughput was measured on NVIDIA V100 single GPU with 128 batch-size.
We trained the network using four representative training environments.
The first is a vanilla setting that trains the network without complicated training techniques.
The vanilla setting has the lowest performance due to the lack of techniques to help generalization performance and also used for the previous experiments in Figure~\ref{fig:resnet_pooling_experiment}, \ref{fig:vit_pooling_experiment}.
The second is training with CutMix~\cite{yun2019cutmix} data augmentation.
Although only data augmentation has changed, it shows significantly better performance than the vanilla setting.
The third is the DeiT~\cite{deit} setting, which is a compilation of training techniques to train ViT on ImageNet-1k~\cite{deng2009imagenet}.
DeiT setting includes various training techniques and parameter tuning, and we used the same training setting through the official open-source code.
However, in the case of Repeated Augment~\cite{hoffer2020repaetedaugment}, we confirmed that it had a negative effect in a small model, and it was used only for Base models.
The last is a DeiT setting with knowledge distillation.
The distillation setting is reported as the best performance setting in DeiT~\cite{deit} paper. 
The network uses an additional distillation token and is trained with distillation loss~\cite{hinton2015distilling} using RegNetY-16GF~\cite{radosavovic2020regnet} as a teacher network.
We used AdamP~\cite{heo2021adamp} optimizer for all settings, and the learning rate, weight decay, and warmup were set equal to DeiT~\cite{deit} paper.
We train models over 100 epochs for Vanilla and CutMix settings, and 300 epochs for DeiT and Distill\alambic settings.

The results are shown in Table~\ref{table:vit_pit_imagenet}.
Comparing the PiT and ViT of the same name, the PiT has fewer FLOPs and faster speed than ViT.
Nevertheless, PiT shows higher performance than ViT.
In the case of vanilla and CutMix settings, where a few training techniques are applied, the performance of PiT is superior to the performance of ViT.
Even in the case of a DeiT and distill settings, PiT shows comparable or better performance to ViT.
Therefore, PiT can be seen as a better architecture than ViT in terms of performance and computation.
The generalization performance issue of ViT in Figure~\ref{fig:vit_pooling_experiment} can also be observed in this experiment.
Like ViT-S in the Vanilla setting and ViT-B in the CutMix setting, ViT often shows no increase in performance even when the model size increases.
On the other hand, the performance of PiT increases according to the model size in all training settings.
it seems that the generalization performance problem of ViT is alleviated by the pooling layers.

We compared the performance of PiT with the convolutional networks.
In the previous experiment, we performed the comparison in the same training setting using the similarity of architecture. 
However, when comparing various architectures, it is infeasible to unify with a setting that works well for all architectures.
Therefore, we performed the comparison based on the best performance reported for each architecture.
But, it was limited to the model trained using only ImageNet images.
When the paper that proposed the architecture and the paper that reported the best performance was different, we cite both papers.
When the architecture is different, the comparison of FLOPs often fails to reflect the actual throughput.
Therefore, we re-measured the GPU throughput and number of params on a single V100 GPU and compared the top-1 accuracy for the performance index.
Table~\ref{table:imagenet_models} shows the comparison result.
In the case of the PiT-B scale, the transformer-based architecture (ViT-B, PiT-B) outperforms the convolutional architecture.
Even in the PiT-S scale, PiT-S shows superior performance than convolutional architecture (ResNet50) or outperforms in throughput (EfficientNet-b3).
However, in the case of PiT-Ti, the performance of convolutional architectures such as ResNet34~\cite{he2016resnet}, MobileNetV3~\cite{howard2019mobilev3}, and EfficientNet-b0~\cite{tan2019efficientnet} outperforms ViT-Ti and PiT-Ti.
Overall, the transformer architecture shows better performance than the convolutional architecture at the scale of ResNet50 or higher, but it is weak at a small scale.
Creating a light-weight transformer architecture such as MobileNet is one of the future works of ViT research.

\begin{table}[t]
\small
\centering
\begin{tabular}{@{}cccc@{}}
\toprule
Setting & Architecture & \begin{tabular}[c]{@{}c@{}}Throughput \\ (imgs/sec)\end{tabular} & Accuracy \\ \midrule
\multirow{7}{*}{\begin{tabular}[c]{@{}c@{}}Long\\training\\(1000 epochs)\end{tabular}} & ViT-Ti\alambic~\cite{deit} & 2564   & 76.6\% \\
& PiT-Ti\alambic             & 3030   & 76.4\% \\
& PiT-XS\alambic             & 2128   & 80.6\% \\ \cmidrule(l){2-4} 
& ViT-S\alambic~\cite{deit} & 980   & 82.6\% \\
& PiT-S\alambic             & 1266   & 82.7\% \\ \cmidrule(l){2-4} 
& ViT-B\alambic~\cite{deit} & 303   & 84.2\% \\
& PiT-B\alambic             & 348   & 84.5\% \\ \midrule
\multirow{4}{*}{\begin{tabular}[c]{@{}c@{}}Large resolution\\(384$\times$384)\end{tabular}} & ViT-B~\cite{deit} & 91   & 83.1\% \\
& PiT-B             & 82   & 83.0\% \\ \cmidrule(l){2-4} 
& ViT-B\alambic~\cite{deit} & 91   & 84.5\% \\
& PiT-B\alambic             & 82   & 84.6\% \\ \bottomrule
\end{tabular}
\caption{\small \textbf{Extended training settings.} We compare the performance of PiT with ViT for long training (1000 epochs) and fine-tune on large resolution (384$\times$384)  }
\label{table:vit_pit_extend_imagenet}
\vspace{-1em}
\end{table}

Additionally, we conduct experiments on two extended training schemes: long training and fine-tune on large resolution. 
Table~\ref{table:vit_pit_extend_imagenet} shows the results. 
As shown in the previous study~\cite{deit}, the performance of ViT is significantly improved on the long training scheme (1000 epochs).
So, we validate PiT on the long training scheme. 
As shown in Table~\ref{table:vit_pit_extend_imagenet}, PiT models show comparable performance with ViT models on the long training scheme.
Although the performance improvement is reduced than the Distill\alambic setting, PiTs still outperform ViT counterparts in throughput.
Fine-tuning on large resolution ($384\times384$) is a famous method to train a large ViT model with small computation.
In the large resolution setting, PiT has comparable performance with ViT, but, worse than ViT on throughput.
It implies that PiT is designed for $224\times224$ and the design is not compatible for the large resolution.
However, we believe that PiT can outperform ViT with a new layer design for $384\times384$.

\subsection{Object detection}

\begin{table}[t]
\small
\centering
\begin{tabular}{@{}c|ccc|cc@{}}
\toprule
\multirow{2}{*}{Backbone} & 
\multicolumn{3}{c|}{Avg. Precision at IOU} &
\multirow{2}{*}{Params.} &
Latency \\
& AP & AP$_{\text{50}}$ & AP$_{\text{75}}$ & & (ms / img) \\ \midrule
ResNet50~\cite{he2016resnet} & 41.5 & 60.5 & 44.3 & 41.0 M  & 49.7 \\
ViT-S~\cite{deit} & 36.9 & 57.0 & 38.0 & 34.9 M & 55.2 \\
PiT-S    & 39.4 & 58.8 & 41.5 & 36.6 M & 46.9 \\
\bottomrule
\end{tabular}
\caption{\textbf{COCO detection performance based on Deformable DETR~\cite{zhu2020deformable}}. We evaluate the performance of PiT as a pretrained backbone for object detection.}
\label{table:detection}
\vspace{-1em}
\end{table}

We validate PiT through object detection on COCO dataset~\cite{lin2014coco} in Deformable-DETR~\cite{zhu2020deformable}.
We train the detectors with different backbones including ResNet50, ViT-S, and our PiT-S.
We follow the training setup of the original paper~\cite{zhu2020deformable} except for the image resolution.
Since the original image resolution is too large for transformer-based backbones, we halve the image resolution for training and test of all backbones.
We use bounding box refinement and a two-stage scheme for the best performance~\cite{zhu2020deformable}.
For multi-scale features for ViT-S, we use features at the 2nd, 8th, and 12th layers following the position of pooling layers on PiT.
All detectors are trained for 50 epochs and the learning rate is dropped by factor 1/10 at 40 epochs.

Table~\ref{table:detection} shows the measured AP score on val2017. 
The detector based on PiT-S outperforms the detector with ViT-S.
It shows that the pooling layer of PiT is effective not only for ImageNet classification but also for pretrained backbone for object detection.
We measured single image latency with a random noise image at resolution $600\times400$
PiT based detector has lower latency than detector based on ResNet50 or ViT-S.
Although PiT detector cannot beat the performance of the ResNet50 detector, PiT detector has better latency, and improvement over ViT-S is significant.
Additional investigation on the training settings for PiT based detectors would improve the performance of the PiT detector.

\subsection{Robustness benchmarks}

\begin{table}[t]
\centering
\small
\setlength{\tabcolsep}{2.5pt}
\begin{tabular}{@{}lccccc@{}}
\toprule
          & Standard & Occ  & IN-A~\cite{hendrycks2019natural} & BGC~\cite{xiao2020noise}  & FGSM~\cite{goodfellow2014explaining} \\ \midrule
PiT-S     & 80.8     & 74.6 & 21.7 & 21.0 & 29.5 \\
ViT-S~\cite{deit}     & 79.8     & 73.0 & 19.1 & 17.6 & 27.2 \\ \midrule
ResNet50~\cite{he2016resnet}  & 76.0     & 52.2 & 0.0  & 22.3 & 7.1  \\ 
ResNet50$^\dagger$~\cite{rw2019timm}  & 79.0     & 67.1 & 5.4  & 32.7 & 24.7  \\ \bottomrule
\end{tabular}
\caption{\textbf{ImageNet robustness benchmarks.} We compare three comparable architectures, PiT-B, ViT-S, and ResNet50 on various ImageNet robustness benchmarks, including center occlusion (Occ), ImageNet-A (IN-A), background challenge (BGC), and fast sign gradient method (FGSM) attack. We evaluate two ResNet50 models from the official PyTorch repository, and the well-optimized implementation~\cite{rw2019timm}, denoted as $\dagger$.}
\label{table:robustness}
\vspace{-1em}
\end{table}

In this subsection, we investigate the effectiveness of the proposed architecture in terms of robustness against input changes.
We presume that the existing ViT design concept, which keeps the spatial dimension from the input layer to the last layer, has two conceptual limitations: {\it Lack of background robustness} and {\it sensitivity to the local discriminative visual features}.
We, therefore, presume that PiT, our new design choice with the pooling mechanism, performs better than ViT for the background robustness benchmarks and the local discriminative sensitivity benchmarks.

We employ four different robustness benchmarks. {\bf Occlusion benchmark} measures the ImageNet validation accuracy where the center $112 \times 112$ patch of the images is zero-ed out. This benchmark measures whether a model only focuses on a small discriminative visual feature or not.
{\bf ImageNet-A} (IN-A) is a dataset constructed by collecting the failure cases of ResNet50 from the web~\cite{hendrycks2019natural} where the collected images contain unusual backgrounds or objects with very small size~\cite{li2021rethinking}. From this benchmark, we can infer how a model is less sensitive to unusual backgrounds or object size changes. However, since IN-A is constructed by collecting images (queried by 200 ImageNet subclasses) where ResNet50 predicts a wrong label, this dataset can be biased towards ResNet50 features.
We, therefore, employ {\bf background challenge} (BGC) benchmark~\cite{xiao2020noise} to explore the explicit background robustness. The BGC dataset consists of two parts, foregrounds, and backgrounds. This benchmark measures the model validation accuracy while keeping the foreground but adversarially changing the background from the other image. Since BGC dataset is built upon nine subclasses of ImageNet, the baseline random chance is 11.1\%.
Lastly, we tested adversarial attack robustness using the fast gradient sign method (FGSM)~\cite{goodfellow2014explaining}.

Table~\ref{table:robustness} shows the results.
First, we observe that PiT shows better performances than ViT in all robustness benchmarks, despite they show comparable performances in the standard ImageNet benchmark (80.8 vs. 79.8). It supports that our dimension design makes the model less sensitive to the backgrounds and the local discriminative features.
Also, we found that the performance drops for occluded samples by ResNet50 are much dramatic than PiT; 80.8 $\rightarrow$ 74.6, 5\% drops for PiT, 79.0 $\rightarrow$ 67.1, 15\% drops for ResNet50. This implies that ResNet50 focuses more on the local discriminative areas, by the nature of convolutional operations.
Interestingly, in Table~\ref{table:robustness}, ResNet50 outperforms ViT variants in the background challenge dataset (32.7 vs. 21.0). This implies that the self-attention mechanism unintentionally attends more backgrounds comparing to ResNet design choice. Overcoming this potential drawback of vision transformers will be an interesting research direction.
\section{Conclusion}

In this paper, we have shown that the design principle widely used in CNNs - the spatial dimensional transformation performed by pooling or convolution with strides, is not considered in transformer-based architectures such as ViT; ultimately affects the model performance. We have first studied with ResNet and found that the transformation in respect of the spatial dimension increases the computational efficiency and the generalization ability. To leverage the benefits in ViT, we propose a PiT that incorporates a pooling layer into Vit, and PiT shows that these advantages can be well harmonized to ViT through extensive experiments. Consequently, while significantly improving the performance of the ViT architecture, we have shown that the pooling layer by considering spatial interaction ratio is essential to a self-attention-based architecture.

{\small
\paragraph{Acknowledgement}
We thank NAVER AI Lab members for valuable discussion and advice. 
NSML~\cite{nsml} has been used for experiments. We thank the reviewers for the productive feedback.
}
{\small
\bibliographystyle{ieee_fullname}
\bibliography{egbib}

\begin{thebibliography}{10}\itemsep=-1pt

\bibitem{ba2016layernorm}
Jimmy~Lei Ba, Jamie~Ryan Kiros, and Geoffrey~E Hinton.
\newblock Layer normalization.
\newblock {\em arXiv preprint arXiv:1607.06450}, 2016.

\bibitem{bello2021lambdanetworks}
Irwan Bello.
\newblock Lambdanetworks: Modeling long-range interactions without attention.
\newblock In {\em International Conference on Learning Representations}, 2021.

\bibitem{cai2018proxylessnas}
Han Cai, Ligeng Zhu, and Song Han.
\newblock Proxylessnas: Direct neural architecture search on target task and
  hardware.
\newblock {\em arXiv preprint arXiv:1812.00332}, 2018.

\bibitem{carion2020detr}
Nicolas Carion, Francisco Massa, Gabriel Synnaeve, Nicolas Usunier, Alexander
  Kirillov, and Sergey Zagoruyko.
\newblock End-to-end object detection with transformers.
\newblock In {\em European Conference on Computer Vision}, pages 213--229.
  Springer, 2020.

\bibitem{cohen2016expressive}
Nadav Cohen, Or Sharir, and Amnon Shashua.
\newblock On the expressive power of deep learning: A tensor analysis.
\newblock In {\em Conference on learning theory}, pages 698--728. PMLR, 2016.

\bibitem{cohen2016inductive}
Nadav Cohen and Amnon Shashua.
\newblock Inductive bias of deep convolutional networks through pooling
  geometry.
\newblock In {\em International Conference on Learning Representations}, 2016.

\bibitem{dai2020funnel}
Zihang Dai, Guokun Lai, Yiming Yang, and Quoc~V Le.
\newblock Funnel-transformer: Filtering out sequential redundancy for efficient
  language processing.
\newblock {\em arXiv preprint arXiv:2006.03236}, 2020.

\bibitem{deng2009imagenet}
Jia Deng, Wei Dong, Richard Socher, Li-Jia Li, Kai Li, and Li Fei-Fei.
\newblock Imagenet: A large-scale hierarchical image database.
\newblock In {\em 2009 IEEE conference on computer vision and pattern
  recognition}, pages 248--255. Ieee, 2009.

\bibitem{vit}
Alexey Dosovitskiy, Lucas Beyer, Alexander Kolesnikov, Dirk Weissenborn,
  Xiaohua Zhai, Thomas Unterthiner, Mostafa Dehghani, Matthias Minderer, Georg
  Heigold, Sylvain Gelly, Jakob Uszkoreit, and Neil Houlsby.
\newblock An image is worth 16x16 words: Transformers for image recognition at
  scale.
\newblock In {\em International Conference on Learning Representations}, 2021.

\bibitem{goodfellow2014explaining}
Ian~J Goodfellow, Jonathon Shlens, and Christian Szegedy.
\newblock Explaining and harnessing adversarial examples.
\newblock {\em arXiv preprint arXiv:1412.6572}, 2014.

\bibitem{han2017pyramid}
Dongyoon Han, Jiwhan Kim, and Junmo Kim.
\newblock Deep pyramidal residual networks.
\newblock In {\em Proceedings of the IEEE conference on computer vision and
  pattern recognition}, pages 5927--5935, 2017.

\bibitem{han2020rexnet}
Dongyoon Han, Sangdoo Yun, Byeongho Heo, and YoungJoon Yoo.
\newblock Rethinking channel dimensions for efficient model design.
\newblock In {\em Proceedings of the IEEE/CVF Conference on Computer Vision and
  Pattern Recognition}, pages 732--741, 2021.

\bibitem{he2016resnet}
Kaiming He, Xiangyu Zhang, Shaoqing Ren, and Jian Sun.
\newblock Deep residual learning for image recognition.
\newblock In {\em Proceedings of the IEEE conference on computer vision and
  pattern recognition}, pages 770--778, 2016.

\bibitem{He_2019_bag}
Tong He, Zhi Zhang, Hang Zhang, Zhongyue Zhang, Junyuan Xie, and Mu Li.
\newblock Bag of tricks for image classification with convolutional neural
  networks.
\newblock In {\em Proceedings of the IEEE/CVF Conference on Computer Vision and
  Pattern Recognition (CVPR)}, June 2019.

\bibitem{hendrycks2019natural}
Dan Hendrycks, Kevin Zhao, Steven Basart, Jacob Steinhardt, and Dawn Song.
\newblock Natural adversarial examples.
\newblock {\em arXiv preprint arXiv:1907.07174}, 2019.

\bibitem{heo2021adamp}
Byeongho Heo, Sanghyuk Chun, Seong~Joon Oh, Dongyoon Han, Sangdoo Yun, Gyuwan
  Kim, Youngjung Uh, and Jung-Woo Ha.
\newblock Adamp: Slowing down the slowdown for momentum optimizers on
  scale-invariant weights.
\newblock In {\em International Conference on Learning Representations}, 2021.

\bibitem{hinton2015distilling}
Geoffrey Hinton, Oriol Vinyals, and Jeff Dean.
\newblock Distilling the knowledge in a neural network.
\newblock {\em arXiv preprint arXiv:1503.02531}, 2015.

\bibitem{hoffer2020repaetedaugment}
Elad Hoffer, Tal Ben-Nun, Itay Hubara, Niv Giladi, Torsten Hoefler, and Daniel
  Soudry.
\newblock Augment your batch: Improving generalization through instance
  repetition.
\newblock In {\em Proceedings of the IEEE/CVF Conference on Computer Vision and
  Pattern Recognition}, pages 8129--8138, 2020.

\bibitem{howard2019mobilev3}
Andrew Howard, Mark Sandler, Grace Chu, Liang-Chieh Chen, Bo Chen, Mingxing
  Tan, Weijun Wang, Yukun Zhu, Ruoming Pang, Vijay Vasudevan, et~al.
\newblock Searching for mobilenetv3.
\newblock In {\em Proceedings of the IEEE/CVF International Conference on
  Computer Vision}, pages 1314--1324, 2019.

\bibitem{nsml}
Hanjoo Kim, Minkyu Kim, Dongjoo Seo, Jinwoong Kim, Heungseok Park, Soeun Park,
  Hyunwoo Jo, KyungHyun Kim, Youngil Yang, Youngkwan Kim, Nako Sung, and
  Jung{-}Woo Ha.
\newblock {NSML:} meet the mlaas platform with a real-world case study.
\newblock {\em CoRR}, abs/1810.09957, 2018.

\bibitem{krizhevsky2012alexnet}
Alex Krizhevsky, Ilya Sutskever, and Geoffrey~E Hinton.
\newblock Imagenet classification with deep convolutional neural networks.
\newblock {\em Advances in neural information processing systems},
  25:1097--1105, 2012.

\bibitem{pono}
Boyi Li, Felix Wu, Kilian~Q Weinberger, and Serge Belongie.
\newblock Positional normalization.
\newblock In H. Wallach, H. Larochelle, A. Beygelzimer, F. d\textquotesingle
  Alch\'{e}-Buc, E. Fox, and R. Garnett, editors, {\em Advances in Neural
  Information Processing Systems}, volume~32. Curran Associates, Inc., 2019.

\bibitem{li2021rethinking}
Xiao Li, Jianmin Li, Ting Dai, Jie Shi, Jun Zhu, and Xiaolin Hu.
\newblock Rethinking natural adversarial examples for classification models.
\newblock {\em arXiv preprint arXiv:2102.11731}, 2021.

\bibitem{lin2014coco}
Tsung-Yi Lin, Michael Maire, Serge Belongie, James Hays, Pietro Perona, Deva
  Ramanan, Piotr Doll{\'a}r, and C~Lawrence Zitnick.
\newblock Microsoft coco: Common objects in context.
\newblock In {\em European conference on computer vision}, pages 740--755.
  Springer, 2014.

\bibitem{liu2018darts}
Hanxiao Liu, Karen Simonyan, and Yiming Yang.
\newblock Darts: Differentiable architecture search.
\newblock {\em arXiv preprint arXiv:1806.09055}, 2018.

\bibitem{radenovic2018gempooling}
Filip Radenovi{\'c}, Giorgos Tolias, and Ond{\v{r}}ej Chum.
\newblock Fine-tuning cnn image retrieval with no human annotation.
\newblock {\em IEEE transactions on pattern analysis and machine intelligence},
  41(7):1655--1668, 2018.

\bibitem{radosavovic2020regnet}
Ilija Radosavovic, Raj~Prateek Kosaraju, Ross Girshick, Kaiming He, and Piotr
  Doll{\'a}r.
\newblock Designing network design spaces.
\newblock In {\em Proceedings of the IEEE/CVF Conference on Computer Vision and
  Pattern Recognition}, pages 10428--10436, 2020.

\bibitem{ramachandran2019stand}
Prajit Ramachandran, Niki Parmar, Ashish Vaswani, Irwan Bello, Anselm Levskaya,
  and Jon Shlens.
\newblock Stand-alone self-attention in vision models.
\newblock In H. Wallach, H. Larochelle, A. Beygelzimer, F. d\textquotesingle
  Alch\'{e}-Buc, E. Fox, and R. Garnett, editors, {\em Advances in Neural
  Information Processing Systems}, volume~32. Curran Associates, Inc., 2019.

\bibitem{sandler2018mobilenetv2}
Mark Sandler, Andrew Howard, Menglong Zhu, Andrey Zhmoginov, and Liang-Chieh
  Chen.
\newblock Mobilenetv2: Inverted residuals and linear bottlenecks.
\newblock In {\em Proceedings of the IEEE conference on computer vision and
  pattern recognition}, pages 4510--4520, 2018.

\bibitem{simonyan2014vgg}
Karen Simonyan and Andrew Zisserman.
\newblock Very deep convolutional networks for large-scale image recognition.
\newblock {\em arXiv preprint arXiv:1409.1556}, 2014.

\bibitem{szegedy2015googlenet}
Christian Szegedy, Wei Liu, Yangqing Jia, Pierre Sermanet, Scott Reed, Dragomir
  Anguelov, Dumitru Erhan, Vincent Vanhoucke, and Andrew Rabinovich.
\newblock Going deeper with convolutions.
\newblock In {\em Proceedings of the IEEE conference on computer vision and
  pattern recognition}, pages 1--9, 2015.

\bibitem{tan2019efficientnet}
Mingxing Tan and Quoc Le.
\newblock Efficientnet: Rethinking model scaling for convolutional neural
  networks.
\newblock In {\em International Conference on Machine Learning}, pages
  6105--6114. PMLR, 2019.

\bibitem{deit}
Hugo Touvron, Matthieu Cord, Matthijs Douze, Francisco Massa, Alexandre
  Sablayrolles, and Herv{\'e} J{\'e}gou.
\newblock Training data-efficient image transformers \& distillation through
  attention.
\newblock In {\em International Conference on Machine Learning}, pages
  10347--10357. PMLR, 2021.

\bibitem{vaswani2017attention}
Ashish Vaswani, Noam Shazeer, Niki Parmar, Jakob Uszkoreit, Llion Jones,
  Aidan~N Gomez, Lukasz Kaiser, and Illia Polosukhin.
\newblock Attention is all you need.
\newblock {\em arXiv preprint arXiv:1706.03762}, 2017.

\bibitem{vig2019analyzing}
Jesse Vig and Yonatan Belinkov.
\newblock Analyzing the structure of attention in a transformer language model.
\newblock {\em arXiv preprint arXiv:1906.04284}, 2019.

\bibitem{wang2020axial}
Huiyu Wang, Yukun Zhu, Bradley Green, Hartwig Adam, Alan Yuille, and
  Liang-Chieh Chen.
\newblock Axial-deeplab: Stand-alone axial-attention for panoptic segmentation.
\newblock In {\em European Conference on Computer Vision}, pages 108--126.
  Springer, 2020.

\bibitem{wang2018nonlocal}
Xiaolong Wang, Ross Girshick, Abhinav Gupta, and Kaiming He.
\newblock Non-local neural networks.
\newblock In {\em Proceedings of the IEEE conference on computer vision and
  pattern recognition}, pages 7794--7803, 2018.

\bibitem{rw2019timm}
Ross Wightman.
\newblock Pytorch image models.
\newblock \url{https://github.com/rwightman/pytorch-image-models}, 2019.

\bibitem{wu2018groupnorm}
Yuxin Wu and Kaiming He.
\newblock Group normalization.
\newblock In {\em Proceedings of the European conference on computer vision
  (ECCV)}, pages 3--19, 2018.

\bibitem{xiao2020noise}
Kai Xiao, Logan Engstrom, Andrew Ilyas, and Aleksander Madry.
\newblock Noise or signal: The role of image backgrounds in object recognition.
\newblock {\em arXiv preprint arXiv:2006.09994}, 2020.

\bibitem{yun2019cutmix}
Sangdoo Yun, Dongyoon Han, Seong~Joon Oh, Sanghyuk Chun, Junsuk Choe, and
  Youngjoon Yoo.
\newblock Cutmix: Regularization strategy to train strong classifiers with
  localizable features.
\newblock In {\em Proceedings of the IEEE/CVF International Conference on
  Computer Vision}, pages 6023--6032, 2019.

\bibitem{yun2021relabel}
Sangdoo Yun, Seong~Joon Oh, Byeongho Heo, Dongyoon Han, Junsuk Choe, and
  Sanghyuk Chun.
\newblock Re-labeling imagenet: from single to multi-labels, from global to
  localized labels.
\newblock In {\em Proceedings of the IEEE/CVF Conference on Computer Vision and
  Pattern Recognition}, pages 2340--2350, 2021.

\bibitem{zhang2020resnest}
Hang Zhang, Chongruo Wu, Zhongyue Zhang, Yi Zhu, Zhi Zhang, Haibin Lin, Yue
  Sun, Tong He, Jonas Mueller, R Manmatha, et~al.
\newblock Resnest: Split-attention networks.
\newblock {\em arXiv preprint arXiv:2004.08955}, 2020.

\bibitem{zhu2020deformable}
Xizhou Zhu, Weijie Su, Lewei Lu, Bin Li, Xiaogang Wang, and Jifeng Dai.
\newblock Deformable detr: Deformable transformers for end-to-end object
  detection.
\newblock {\em arXiv preprint arXiv:2010.04159}, 2020.

\end{thebibliography}
}

\end{document}